\documentclass[11pt]{article}

\usepackage[final]{acl}

\usepackage{times}
\usepackage{latexsym}
\usepackage[dvipsnames]{xcolor}
\colorlet{red}{black}
\usepackage{tabularx}
\usepackage{booktabs}
\usepackage{array}
\let\LaTeXBegin\begin
\let\LaTeXEnd\end

\usepackage{arabtex}
\usepackage{utf8}
\setcode{utf8}

\let\begin\LaTeXBegin
\let\end\LaTeXEnd
\def\endabstract{\unskip\end{list}}

\usepackage{lipsum}
\usepackage{arydshln}

\usepackage[utf8]{inputenc}
\usepackage[T1]{fontenc}

\usepackage{microtype}

\usepackage{graphicx}
\usepackage{enumitem}
\usepackage[normalem]{ulem} 
\usepackage{xcolor} 
\usepackage{url}
\usepackage[most]{tcolorbox}
\usepackage{fancyvrb}
\usepackage{pifont} 

\usepackage{soul} 

\title{StanceEval 2026: \\The Second Stance Detection Shared Task}

\author{
\textbf{Rasha Albalawi}$^{1}$
~ \textbf{Nuha Albadi}$^{1,2}$
~ \textbf{Hamzah Luqman}$^{1,2}$
~ \textbf{Asma Yamani}$^{1}$
~ \textbf{Maram Kurdi}$^{1}$ \\
\textbf{Saad Ezzini}$^{1}$
~ \textbf{Ahmed Ashraf}$^{1}$
~ \textbf{Maged  Al-Shaibani}$^{2}$
~ \textbf{Nora Alturayeif}$^{3}$ \\
$^{1}$Information and Computer Science Department, KFUPM, Saudi Arabia \\
$^{2}$SDAIA--KFUPM Joint Research Center for Artificial Intelligence, Saudi Arabia \\
$^{3}$HUMAIN, Saudi Arabia \\
\texttt{\{g202521490, nuha.badi, hluqman, g201906630, maram.kurdi,} \\
\texttt{saad.ezzini, g202411740, majed.alshaibani\}@kfupm.edu.sa} \\
\texttt{nalturayeif@humain.com}
}

\begin{document}
\maketitle

\begin{abstract}

StanceEval 2026\footnote{\href{https://stanceeval.github.io/}{StanceEval 2026 website: https://stanceeval.github.io/}}  is the second edition of the StanceEval shared task series on stance detection in Arabic social media text. Stance detection aims to identify a writer's stance toward a given topic. Given a tweet and a target, participating systems must determine whether the writer's stance is \textit{Favor}, \textit{Against}, or \textit{None}. This edition focuses on cross-target generalization across two distinct evaluation tracks: Track~1 evaluates thematically related cross-target transfer (testing on \textit{Women Driving}, related to \textit{Women Empowerment} from training data), while Track~2 evaluates cross-domain transfer to completely unseen targets (\textit{E-Cars} and \textit{Trimester System}). The shared task attracted 80 registered teams from 12 countries. During the evaluation phase, 30 unique teams submitted entries, with 21 teams officially ranked in Track~1 and 13 in Track~2 following validation filtering, and 20 teams submitting system-description papers. Participating teams employed diverse methodologies, including fine-tuned pretrained language models, prompt-based and retrieval-augmented large language models (LLMs), fine-tuned LLMs, and hybrid cascades. Top systems achieved impressive $F_{avg2}$ scores of 0.8994 on Track~1 and 0.9400 on Track~2, substantially outperforming the strongest baselines (0.7366 and 0.7475, respectively)\textcolor{red}{, where $F_{avg2}$ denotes the macro-averaged F1 score over the \textit{Favor} and \textit{Against} classes}. Counterintuitively, performance on the unseen targets was higher than on the related target, a disparity could be driven by extreme target polarization, class imbalance, and dialectal or sarcastic nuance across topics.

\end{abstract}
\section{Introduction}

Social media platforms contain large volumes of user-generated content expressing positions on political, social, and public issues. Arabic stance detection is particularly challenging due to dialectal variation, informal expressions, spelling variations, and implicit language. Moreover, many existing stance detection benchmarks train and evaluate systems on the same predefined targets, providing limited evidence of their ability to generalize to new topics.

StanceEval 2024 introduced the first shared task for Arabic stance detection~\cite{alturayeif2024stanceeval}, using the Mawqif dataset~\cite{alturayeif2022mawqif} of 4,121 Arabic tweets covering COVID-19 Vaccine, Digital Transformation, and Women Empowerment. Each tweet is annotated for stance, sentiment, and sarcasm. However, training and evaluation covered the same targets, leaving cross-target generalization unexplored.

StanceEval 2026 addresses this limitation by focusing on cross-target generalization. It uses Mawqif-XT~\cite{albalawi2026mawqif}, which adds 996 manually annotated Arabic tweets covering Women Driving, E-Cars, and Trimester System. Track 1 evaluates related-target generalization on Women Driving, which is unseen during training but semantically related to Women Empowerment. Track 2 evaluates unseen-target generalization on E-Cars and Trimester System, whose domains are absent from the training and development data. Systems classify each tweet as Favor, Against, or None toward its target. Sentiment and sarcasm annotations are also provided to support stance prediction.

The shared task received 80 registrations from 12 countries and 30 teams participated in the evaluation, resulting in 21 and 13 officially ranked teams in Track1 and Track2, respectively, after filtering invalid or non-compliant submissions, and 20 accepted system description papers. Systems used approaches including prompting-based and retrieval-augmented large language models (LLMs), fine-tuned Arabic encoders, multi-task learning, and hybrid cascades. The highest $F_{avg2}$ scores were 0.8994 and 0.9400 for Track1 and Track2, exceeding the best baseline scores of 0.7366 and 0.7475, respectively.

This paper presents an overview of StanceEval 2026, describing the task design, datasets, evaluation, baselines, results, and participating systems, followed by a discussion of key findings and limitations.
\section{Literature Review}\label{sec:lit}

Stance detection methods have evolved from feature-based classifiers to neural architectures, pretrained language models, and, more recently, LLMs. Early approaches relied on supervised classifiers such as SVMs \cite{lai2019stance}, followed by neural models including CNNs \cite{zhou2019condensed} and RNNs \cite{yang2020tweet}. Given the scarcity of labeled stance data, transfer learning and fine-tuning pretrained models became widely adopted \cite{alturayeif2023b}. More recently, LLMs have been explored as an alternative to fine-tuning \cite{pangtey2025llmstance}. \newcite{cruickshank2024prompting} showed that zero- and few-shot prompting can match supervised baselines on several stance detection benchmarks, although performance varies across targets and datasets.

This variation across targets reflects a broader challenge: stance detection systems are typically evaluated on fixed target sets and may not transfer well to unseen targets. This has motivated research on cross-target and zero-shot stance detection. \newcite{xu2018cross} modeled relationships between source and destination targets using self-attention, while \newcite{allaway2020zeroshot} and \newcite{allaway2021adversarial} proposed generalized topic representations and adversarial training, respectively, for unseen targets. The Will-They-Won’t-They corpus \cite{conforti2020willtheywontthey} was developed to stress-test this setting at scale. More recently, \newcite{gera2025mlsd} used few-shot metric learning to improve transfer across targets and domains, while \newcite{li2026zeroshot} used an LLM to generate and adapt to candidate targets dynamically rather than assuming test targets are known in advance. Nevertheless, performance can drop sharply when test targets differ from training targets in topic, framing, or dialect \cite{allaway2023paradigms, khiabani2025crosstarget}.

Arabic stance detection has received less attention than English, with existing resources largely covering fixed sets of targets. \newcite{alturayeif2022mawqif} released MAWQIF, the resource used in StanceEval 2024 \cite{alturayeif2024stanceeval}, while \newcite{charfi2024marasta} introduced MARASTA, a cross-domain, multi-dialectal Arabic stance corpus. More recently, StanceNakba \cite{aldous2026stancenakba} introduced a cross-topic stance detection subtask on Arabic conflict-related discourse, reflecting growing interest in evaluating generalization beyond fixed targets. StanceEval 2026 continues this direction with Mawqif-XT \cite{albalawi2026mawqif}, which introduces targets held out from training to evaluate systems on  related and unrelated topics.


\section{Task Description and Dataset}\label{sec:task_description}
StanceEval 2026 focuses on cross-target generalization in Arabic stance detection. Given an Arabic tweet and a target, participating systems classify the stance as \textit{Favor}, \textit{Against}, or \textit{None}. The shared task comprises two tracks that evaluate generalization to related and unseen targets.

The three labels follow the annotation scheme of the Mawqif dataset \cite{alturayeif2022mawqif}. \textit{Favor} indicates explicit or implicit support for the target, while \textit{Against} indicates explicit or implicit opposition. \textit{None} is assigned when the tweet does not express a clear stance toward the target or is irrelevant to it.

StanceEval 2026 uses the original Mawqif dataset
\cite{alturayeif2022mawqif} for training and development and Mawqif-XT \cite{albalawi2026mawqif} for testing. Table~\ref{tab:target_stats} summarizes the number of tweets and stance distribution across the six targets. \textcolor{red}{Representative examples from the datasets, including their corresponding labels, are provided in Appendix~\ref{app:examples}.} Mawqif-XT is publicly released, with  data availability details provided in \cite{albalawi2026mawqif}.

Tweets in both datasets were collected from Twitter (now X) using
target-specific Arabic keywords and hashtags. Mawqif-XT used additional target-oriented filtering and was manually annotated by three annotators for stance, sarcasm, and sentiment, with adjudication used to establish the final gold labels. The data contain naturally occurring Arabic social-media discourse, including dialectal expressions, informal language, and spelling variation. Further details on data collection, annotation, agreement, and quality control are provided in \cite{alturayeif2022mawqif,albalawi2026mawqif}.

The two tasks evaluate cross-target generalization under related-target
and unseen-target settings, with their respective data configurations
summarized in Table~\ref{tab:protocols}.

\begin{table}[t]
\centering

\resizebox{\columnwidth}{!}{%
\begin{tabular}{lcccc}
\toprule
\textbf{Target} &
\textbf{Tweets} &
\textbf{Favor (\%)} &
\textbf{Against (\%)} &
\textbf{None (\%)} \\
\midrule

\multicolumn{5}{l}{\textbf{Mawqif}} \\
\midrule
COVID-19 Vaccine
& 1,373 & 43.63 & 43.55 & 12.82 \\

Digital Transformation
& 1,348 & 76.78 & 12.39 & 10.83 \\

Women Empowerment
& 1,400 & 63.86 & 31.14 & 5.00 \\

\midrule

\multicolumn{5}{l}{\textbf{Mawqif-XT}} \\
\midrule
Women Driving
& 352 & 44.89 & 45.45 & 9.66 \\

Trimester System
& 312 & 22.76 & 70.19 & 7.05 \\

E-Cars
& 332 & 43.67 & 35.24 & 21.08 \\

\bottomrule
\end{tabular}%
}
\caption{\textcolor{red}{Target statistics for StanceEval 2026.}}\label{tab:target_stats}
\end{table}

\begin{table*}[t]
\centering
\small
\renewcommand{\arraystretch}{1.25}

\begin{tabular}{p{1.5cm} p{2.5cm} p{2.5cm} p{3.8cm} p{3.8cm}}
\toprule
&
\multicolumn{2}{c}{\textbf{Mawqif}} &
\multicolumn{1}{c}{\textbf{Mawqif-XT}} &
\\
\cmidrule(lr){2-3}
\cmidrule(lr){4-4}

\textbf{\textcolor{red}{Track}} &
\textbf{Training} &
\textbf{Development} &
\textbf{Test} &
\textbf{Evaluation Setting} \\
\midrule

\textcolor{red}{Track 1}  &
\shortstack[l]{CV, DT, WE\\3,502 tweets} &
\shortstack[l]{CV, DT, WE\\619 tweets} &
\shortstack[l]{Women Driving\\352 tweets} &
Related-target generalization \\

\midrule

\textcolor{red}{Track 2} &
\shortstack[l]{CV, DT\\\textcolor{red}{2,721} tweets} &
\shortstack[l]{WE\\ \textcolor{red}{1400}  tweets} &
\shortstack[l]{Trimester System, E-Cars\\644 tweets} &
Unseen-target generalization \\

\bottomrule
\end{tabular}

\vspace{1mm}
\caption{\textcolor{red}{Training, development, and test setup for StanceEval 2026.}}
\label{tab:protocols}
\footnotesize
CV = COVID-19 Vaccine, DT = Digital Transformation, and WE = Women Empowerment.

\end{table*}

\paragraph{Evaluation Metric.}
Both tasks use $F_{avg2}$, $F_{avg3}$, and overall accuracy.
$F_{avg2}$ is the macro-average F1 over \textit{Favor} and
\textit{Against} and serves as the primary ranking metric, while
$F_{avg3}$ is the macro-average F1 over all three stance labels.

\subsection*{Track 1: Thematically Related Cross-Target Stance Detection} 

\paragraph{Task Description.} 
Track~1 evaluates whether stance knowledge learned from existing targets can transfer to a new, previously unobserved target within a semantically related thematic domain. Participants are required to predict the stance of Arabic tweets toward \textit{Women Driving}. While the target itself is not included in the training or development data, it is conceptually and culturally related to \textit{Women Empowerment}, one of the targets available during model development. This setting therefore evaluates the ability of stance detection systems to transfer knowledge from a previously observed target to a new target within a shared social domain. 

\paragraph{Data.}
The training and development sets for Track~1 are taken from the original Mawqif dataset \cite{alturayeif2022mawqif}. Both sets contain tweets associated with the three original targets: \textit{COVID-19 Vaccine}, \textit{Digital Transformation}, and \textit{Women Empowerment}. The test set is drawn from Mawqif-XT \cite{albalawi2026mawqif} and contains 352 tweets associated with \textit{Women Driving}. 

\subsection*{Track 2: Thematically Unseen Cross-Target Stance Detection} 
Track~2 evaluates a broader cross-domain stance setting in which systems are tested on targets that are completely absent from both the training and development sets and belong to distinct topical domains. Participants are asked to predict the stance of Arabic tweets toward two new targets: \textit{E-Cars} and \textit{Trimester System}. These targets cover different domains (automotive technology/infrastructure and education reform) from the targets used for model development. The task examines how well stance detection models can generalize out-of-domain without target-specific fine-tuning data.

\paragraph{Data.}
The training and development data for Track~2 are taken from the original Mawqif dataset \cite{alturayeif2022mawqif}. The training set contains tweets from \textit{COVID-19 Vaccine} and \textit{Digital Transformation}, while \textit{Women Empowerment} is reserved for development. The test set is drawn from Mawqif-XT \cite{albalawi2026mawqif} and consists of 332 tweets for \textit{E-Cars} and 312 tweets for \textit{Trimester System}.

\section{Shared Task Teams \& Results}\label{sec:teams_results}
\subsection{Participating Teams}
We received 80 registration responses for StanceEval 2026. During the evaluation phase, 30 unique teams submitted entries to at least one of the two tracks. Of these, 11 teams (36.7\%) submitted only to Track~1, 2 teams (6.7\%) submitted only to Track~2, and 17 teams (56.7\%) participated in both tracks. Before finalizing the official standings, submissions were audited to ensure compliance with task guidelines: entries from unregistered participants, systems utilizing manual test-set annotations, and repeated probing attempts designed to infer ground-truth test labels were excluded. Following this review, 21 teams were officially ranked in Track~1 and 13 teams in Track~2, with 20 teams submitting accepted system-description papers.

The participating teams represented universities, research laboratories, industry organizations, and independent researchers across 12 countries. Affiliation information could not be established for five teams (16.7\%), while the remaining 25 teams (83.3\%) represented a range of academic (46.7\%), industry (13.3\%), collaborative (23.3\%), and independent backgrounds. Saudi Arabia contributed the largest contingent of teams, followed by India, Pakistan, Bangladesh, Qatar, Vietnam, and Egypt, each accounting for 6.7\%–16.7\% of the participating teams, with the United States, Iran, Morocco, the United Kingdom, and Germany each contributing one team.

\subsection{Baselines}

We provide a set of baseline systems as reference points for comparison with the participating systems. The baselines include Arabic and multilingual pretrained language models, as well as LLMs evaluated in a zero-shot setting.

\subsubsection{Experimental Setup}
\paragraph{Models}
We evaluated four Arabic pretrained language models: AraBERT-v02~\cite{antoun2020arabert}, AraBERT-twitter~\cite{antoun2020arabert}, CAMeLBERT-da~\cite{inoue2021interplay}, and MARBERT~\cite{abdul2021arbert}. We also evaluated three multilingual models: mBERT~\cite{devlin2019bert}, DistilBERT~\cite{sanh2019distilbert}, and XLM-RoBERTa~\cite{conneau2020unsupervised}. In addition, three LLMs were evaluated in a zero-shot setting: LLaMA 3.3 (70B)~\cite{touvron2023llama}, Qwen 2.5 (72B)~\cite{qwen2025qwen25technicalreport}, and JAIS (70B)~\cite{sengupta2023jais}.

\paragraph{Training Configuration.}
All encoder-based models were trained using the same settings. The target and tweet were provided as a sentence pair, with a maximum sequence length of 128 and a batch size of 32. We fine-tuned the models for 20 epochs using AdamW with a learning rate of $2\times10^{-5}$ and selected the best checkpoint based on development set performance. For the zero-shot LLMs, we provided the target and tweet as input and the models \textcolor{red}{were} instructed to predict one of the three stance labels: Favor, Against, or None. \textcolor{red}{The same instruction prompt was used for all models, with no demonstration examples. We used 4-bit quantization, a maximum input length of 384 tokens, a maximum of 2 generated tokens, a batch size of 2, and deterministic decoding with a temperature of 0.0.} \textcolor{red}{The exact zero-shot prompt is provided in Appendix~\ref{app:llm_prompt}.}

\paragraph{Evaluation.}
The baseline experiments follow the two StanceEval-2026 tracks. In Track~1, models are trained and validated on the original Mawqif targets and evaluated on the Women Driving target from Mawqif-XT. In Track~2, models are trained on COVID-19 Vaccine and Digital Transformation, validated on Women Empowerment, and evaluated on E-Cars and Trimester System targets from Mawqif-XT. The primary evaluation metric is $F_{avg2}$, computed as the macro-average F1 score over the Favor and Against classes.

\subsubsection{Baseline Results}

Table~\ref{tab:baseline-test-results} summarizes the baseline performance on the two StanceEval 2026 tracks using $F_{\mathrm{avg2}}$, $F_{\mathrm{avg3}}$, and accuracy. In Track~1, Qwen 2.5 achieves the highest $F_{\mathrm{avg2}}$ score (\textcolor{red}{0.7366}), while JAIS obtains the best $F_{\mathrm{avg3}}$ (\textcolor{red}{0.5714}) and accuracy (\textcolor{red}{0.6989}). In Track~2, Qwen 2.5 achieves the highest $F_{\mathrm{avg2}}$ (\textcolor{red}{0.7475}) and $F_{\mathrm{avg3}}$ (\textcolor{red}{0.6101}), whereas MARBERT attains the highest accuracy (\textcolor{red}{0.6910}). Overall, the results show that the strongest model depends on the evaluation metric, although zero-shot LLMs achieve the best $F_{\mathrm{avg2}}$ performance in both tracks.
\begin{table*}[htbp]
\centering
\scriptsize

\begin{tabular}{@{}lccc|ccc@{}}
\toprule
& \multicolumn{3}{c|}{\textbf{Track 1: Related Target (WD)}}
& \multicolumn{3}{c}{\textbf{Track 2: Unseen Targets (EC + TR)}} \\

\cmidrule(lr){2-4}
\cmidrule(lr){5-7}

\textbf{Model}
& $F_{\mathrm{avg2}}$
& $F_{\mathrm{avg3}}$
& Accuracy
& $F_{\mathrm{avg2}}$
& $F_{\mathrm{avg3}}$
& Accuracy \\

\midrule

\multicolumn{7}{@{}l}{\textit{Arabic Encoders}} \\

CAMeLBERT-da
& \textcolor{red}{0.6249}
& \textcolor{red}{0.4166}
& \textcolor{red}{0.5938}
& \textcolor{red}{0.6265}
& \textcolor{red}{0.4923}
& \textcolor{red}{0.5963} \\

MARBERT
& \textcolor{red}{0.6453}
& \textcolor{red}{0.4554}
& \textcolor{red}{0.6023}
& \textcolor{red}{0.7221}
& \textcolor{red}{0.5916}
& \textbf{\textcolor{red}{0.6910}} \\

AraBERT-v02
& \textcolor{red}{0.6592}
& \textcolor{red}{0.4877}
& \textcolor{red}{0.6108}
& \textcolor{red}{0.6782}
& \textcolor{red}{0.5751}
& \textcolor{red}{0.6366} \\

AraBERT-twitter
& \textcolor{red}{0.6484}
& \textcolor{red}{0.4640}
& \textcolor{red}{0.5824}
& \textcolor{red}{0.6277}
& \textcolor{red}{0.5552}
& \textcolor{red}{0.5807} \\

\midrule

\multicolumn{7}{@{}l}{\textit{Multilingual Encoders}} \\

mBERT
& \textcolor{red}{0.6169}
& \textcolor{red}{0.4279}
& \textcolor{red}{0.5938}
& \textcolor{red}{0.5418}
& \textcolor{red}{0.4588}
& \textcolor{red}{0.5140} \\

DistilBERT-multilingual
& \textcolor{red}{0.5643}
& \textcolor{red}{0.4095}
& \textcolor{red}{0.5199}
& \textcolor{red}{0.5521}
& \textcolor{red}{0.4590}
& \textcolor{red}{0.5264} \\

XLM-R-base
& \textcolor{red}{0.6039}
& \textcolor{red}{0.4784}
& \textcolor{red}{0.5568}
& \textcolor{red}{0.5649}
& \textcolor{red}{0.4999}
& \textcolor{red}{0.5388} \\

\midrule

\multicolumn{7}{@{}l}{\textit{Zero-shot LLMs}} \\

LLaMA 3.3 (70B)
& \textcolor{red}{0.6708}
& \textcolor{red}{0.4924}
& \textcolor{red}{0.6307}
& \textcolor{red}{0.6639}
& \textcolor{red}{0.5755}
& \textcolor{red}{0.6289} \\

Qwen 2.5 (72B)
& \textbf{\textcolor{red}{0.7366}}
& \textcolor{red}{0.5444}
& \textcolor{red}{0.6960}
& \textbf{\textcolor{red}{0.7475}}
& \textbf{\textcolor{red}{0.6101}}
& \textcolor{red}{0.6475} \\

JAIS (70B)
& \textcolor{red}{0.7267}
& \textbf{\textcolor{red}{0.5714}}
& \textbf{\textcolor{red}{0.6989}}
& \textcolor{red}{0.6910}
& \textcolor{red}{0.5940}
& \textcolor{red}{0.6180} \\

\bottomrule
\end{tabular}

\caption{\textcolor{red}{Baseline performance on StanceEval 2026.}}
\label{tab:baseline-test-results}

\end{table*}

\subsection{Results}
\label{sec:results}
In this subsection, we present the evaluation results for both Track 1 and Track 2 of the shared task. The leaderboards are sorted based on $F_{avg2}$ (macro-F1 score over the Favor and Against classes). We additionally report $F_{avg3}$ (macro-F1 over all three classes, including None) and overall accuracy for each system. Table~\ref{tab:results_top5} summarizes the five highest-ranked systems in each track; the complete leaderboards are given in Appendix~\ref{app:leaderboards}.


\begin{table}[h]
\centering
\small

\begin{tabular}{c l r r r}
\toprule
\# & Team & $F_{avg2}$ & $F_{avg3}$ & Acc. \\
\midrule
\multicolumn{5}{l}{\textit{Track 1 (related target)}} \\
1 & JLShen & 0.8994 & 0.6646 & 0.8580 \\
2 & ZHR & 0.8960 & 0.6862 & 0.8551 \\
3 & NAMAA & 0.8719 & 0.6722 & 0.8153 \\
4 & Saigum & 0.8685 & 0.6457 & 0.8210 \\
5 & HardStance & 0.8661 & 0.6746 & 0.8267 \\
\midrule
\multicolumn{5}{l}{\textit{Track 2 (unseen targets)}} \\
1 & ZHR & 0.9400 & 0.8792 & 0.9224 \\
2 & JLShen & 0.9357 & 0.8670 & 0.9130 \\
3 & SRK AI Lab & 0.9192 & 0.8374 & 0.8882 \\
4 & Thakaa & 0.9178 & 0.8353 & 0.8882 \\
5 & AH-Translit & 0.9125 & 0.8360 & 0.8820 \\
\bottomrule
\end{tabular}
\caption{\textcolor{red}{Top five systems in each track on the Evaluation set, ranked by $F_{avg2}$. Full leaderboards are provided in Tables~\ref{tab:results_track_1} and~\ref{tab:results_track_2}.}}
\label{tab:results_top5}
\end{table}

\subsubsection{Track 1 Results}
In Track 1, systems were evaluated on the related target, \textit{Women Driving}. As shown in Table~\ref{tab:results_top5}, the leading system, submitted by \textbf{JLShen} team, achieved the highest $F_{avg2}$ score of 0.8994, followed by \textbf{ZHR} with an $F_{avg2}$ of 0.8960, while \textbf{NAMAA} placed third with 0.8719, demonstrating highly competitive results at the top of the leaderboard. Of the 21 ranked teams (Table~\ref{tab:results_track_1}), fourteen exceeded the strongest baseline reported in Table~\ref{tab:baseline-test-results} (Qwen 2.5, 0.7366), and the winning system improved upon it by 0.163 points.
\textcolor{red}{The top-ranked systems achieved closely comparable performance, while $F_{avg3}$ remained consistently lower than $F_{avg2}$, indicating that the \textit{None} class remained challenging even on the related target.}

\subsubsection{Track 2 Results}
Track 2 included 13 participating teams and evaluated systems on the two unseen targets of \textit{E-Cars} and \textit{Trimester System}. The competition in this track was notably close among the top participants (Table~\ref{tab:results_top5}). Team \textbf{ZHR} secured the highest with an $F_{avg2}$ score of 0.9400. Teams \textbf{JLShen} and \textbf{SRK AI Lab} followed closely with 0.9357 and 0.9192 respectively, and the seven highest-ranked systems all achieved $F_{avg2}$ scores above 0.9000 within a margin of 0.038, underscoring the strong and robust performance of these models. Eleven of the 13 ranked teams (Table~\ref{tab:results_track_2}) exceeded the strongest baseline (Qwen 2.5, 0.7475), and the winning system improved upon it by 0.193.
\textcolor{red}{The leading systems remained strong across the reported metrics, with \textbf{ZHR} and \textbf{JLShen} among the top performers. Compared with Track~1, the smaller difference between $F_{avg2}$ and $F_{avg3}$ suggests that the \textit{None} class was less influential in differentiating the strongest systems on the unseen targets.}

\subsection{Discussion: Cross-Target Performance}
A central and counterintuitive empirical finding in StanceEval 2026 is that systems consistently achieved higher performance on Track~2 (thematically unseen targets) than on Track~1 (thematically related target). In Track~2, the winning system achieved an $F_{avg2}$ of 0.9400 (with seven teams scoring above 0.9000 and a mean of 0.861), whereas the winning system in Track~1 achieved an $F_{avg2}$ of 0.8994 (with a mean of 0.788).  
This pattern was also evident among the baseline models: Qwen 2.5 achieved 
\textcolor{red}{0.7475} on Track~2 versus \textcolor{red}{0.7366} on Track~1, 
and MARBERT reached \textcolor{red}{0.7221} on Track~2 compared to 
\textcolor{red}{0.6453} on Track~1.

A conventional view of cross-target generalization suggests that knowledge transfers more effectively between semantically related targets than to completely unseen, out-of-domain targets. However, \textcolor{red}{three factors may help explain this difference, but the current evaluation does not allow us to confirm their individual effects:}

\paragraph{Class Imbalance and Stance Polarization.}
As detailed in Table~\ref{tab:target_stats}, the test targets exhibit markedly different class distributions. The \textit{Trimester System} target in Track~2 is intensely polarized, with $70.19\%$ of tweets expressing opposition (\textit{Against}), $22.76\%$ \textit{Favor}, and only $7.05\%$ \textit{None}. The online discourse surrounding this educational reform expressed sharp, direct public dissatisfaction,  \textcolor{red}{which may have made opposition easier for models, especially instruction-tuned LLMs and retrieval-augmented systems, to identify through explicit lexical and sentiment cues.} In contrast, the \textit{Women Driving} target in Track~1 represents a balanced, highly contested public debate ($44.89\%$ \textit{Favor} vs. $45.45\%$ \textit{Against}). In such balanced discourse, emotional polarity and stance frequently diverge, requiring systems to resolve complex pragmatic nuances.

\paragraph{Rhetorical Nuance, Sarcasm, and Conditional Stances.}
Public discussion on \textit{Women Driving} frequently featured sarcasm, religious and legal argumentation, and conditional support (e.g., agreeing with the reform under strict socio-cultural caveats). Stance towards societal and gender reforms in Arabic social media relies heavily on cultural presuppositions that are difficult for zero-shot models and standard encoders to capture without target-specific supervision. Conversely, discourse on \textit{E-Cars} ($43.67\%$ \textit{Favor}, $35.24\%$ \textit{Against}, $21.08\%$ \textit{None}) centered predominantly on concrete technical, environmental, and financial factors (e.g., charging infrastructure, battery life, cost), where positive and negative stances align more directly with explicit sentiment polarity.

\paragraph{Target Lexical Specificity and the NONE Class Penalty.}
Because the primary metric $F_{avg2}$ evaluates macro-F1 over \textit{Favor} and \textit{Against} alone, the prevalence of the \textit{None} class \textcolor{red}{can affect} system rankings. On targets where \textit{None} represents a small fraction of the data (such as \textit{Trimester System}, $7.05\%$), the risk of confusing neutral text with a stance class is relatively low. For \textit{Women Driving} ($9.66\%$) and \textit{E-Cars} ($21.08\%$), the boundary between informational/neutral commentary and actual author stance is considerably blurred. \textcolor{red}{This difference in class distribution may have reduced stance--neutral confusion for some Track~2 targets and contributed to the higher $F_{avg2}$ scores observed on Track~2.}
\subsection{Analysis of Participating Systems}\label{subsec:system_descriptions}
This section summarizes the submitted systems, grouped by their dominant methodological approach: encoder fine-tuning, prompt-based LLM methods, hybrid encoder-LLM pipelines, fine-tuned LLM ensembles, and LLM-generated reasoning used as an auxiliary signal. Full descriptions of every submitted system, including those not detailed here, appear in Table~\ref{tab:system_synthesis} and Appendix~\ref{app:systems}.

\paragraph{Encoder-Based Fine-Tuning Methods.}
These teams built their systems primarily around fine-tuning pretrained Arabic encoder models for stance classification (TTLab and Nabra are detailed in Appendix~\ref{app:systems}).

Razan Alsulieman~\cite{alsulieman-2026} focuses on validation and target transfer rather than solely on system performance. 
The core system is based on MARBERT with a target-post formulation, extended with ensembling, threshold optimization, NLI/PET formulations, multitask learning, and target adaptation. For Track 2, the authors use a two-way Leave-One-Target-Out validation strategy with paired random seeds, requiring consistent improvement across both directions before accepting a configuration. They conclude that same-target validation substantially overestimated transfer performance for Track 1, whereas held-target validation gave a more reliable estimate under target shift for Track 2.

\paragraph{Prompt-Based LLM Methods.}
Several teams used LLMs with prompting instead of task-specific fine-tuning, including model voting, prompt engineering, and few-shot examples (\textcolor{red}{HardStance} is detailed in Appendix~\ref{app:systems}).



SRK AI Lab~\cite{srk-2026} proposed a multi-LLM system for both tracks. \textcolor{red}{Multiple LLMs independently classify each post, after which GPT-5.5 reviews their predictions and the final label is selected using the predictions, audit observations, and evidence from the target and post.}

JLShen~\cite{jlshen-2026} proposed prompting-based systems for both tracks. \textcolor{red}{Track~1 used a six-LLM ensemble under zero- and few-shot settings with cross-provider voting, while Track~2 used a zero-shot LLM with target-specific Arabic cards followed by selective prediction review. The authors also reconstructed confusion matrices from leaderboard metrics, identifying gold-\textit{NONE} instances misclassified as stance as the main error pattern and using this to inform post-processing, especially for Track~1.}

\paragraph{Hybrid Encoder–LLM Methods}
These teams combined Arabic encoder models with LLMs, differing mainly in how the LLM was used, for routing uncertain cases, fallback on low-confidence predictions, pseudo-label generation, or direct comparison against encoder baselines (U4RASD, Orni, \textcolor{red}{saliyah}, phonghoang, Axiom, and \textcolor{red}{Meta\_Labs} are detailed in Appendix~\ref{app:systems}).


\textcolor{red}{NAMAA Community~\cite{namaa-2026} ensembles Gemini 3.1 Pro, Gemma 4 31B, and GPT-4o using a data-centric Retrieval-Augmented In-Context Learning approach. Each backend retrieves label-balanced demonstrations from cleaned text, adapts to hashtags and negation, and is given injected target descriptions.}

\paragraph{Fine-tuned LLM Ensembles.}
Two submissions built their systems around fine-tuning open LLMs and combining them into ensembles rather than relying purely on prompting (\textcolor{red}{rabia\_naz} is detailed in Appendix~\ref{app:systems}).

ZHR~\cite{zhr-2026} treats both StanceEval tracks as one cross-target problem. The system blends a retrieval-augmented few-shot classifier, in which a fine-tuned MARBERTv2 embedder retrieves demonstrations for an instruction-tuned LLM (Gemma-4-31B-IT) to answer from, with LoRA-adapted generative models (ALLaM-7B-Instruct and Qwen2.5-7B-Instruct).

\paragraph{LLM-Generated Reasoning as Auxiliary Signal.}
These teams used LLM-generated reasoning, such as arguments, discourse annotations, or target descriptions, as an auxiliary signal feeding a downstream classifier, retriever, or scoring step, rather than as the final stance decision itself (Tarjih is detailed in Appendix~\ref{app:systems}).

Saigum~\cite{saigum-2026} combines LLM prompting with retrieval over AI-generated discourse annotations. The system prompts the same LLM $10$ times per tweet for stance, then takes the majority vote. Any tweet where the winning vote share falls below 80\%, or where the vote disagrees with a separate baseline model, gets flagged for a second pass. Each flagged tweet is tagged with seven AI-generated discourse features and a similarity score over these tags retrieves the most relevant labeled examples, which are shown to the model as few-shot demonstrations to produce a revised decision.

Thakaa~\cite{thakaa-2026} supplies target-specific evidence at inference time instead of adapting model parameters, using different strategies per track: for Track 1, it retrieves the eight most similar labeled examples from an external corpus (ArabicStanceX~\cite{alkhathlan-2025-arabicstancex}) via TF-IDF cosine similarity as few-shot demonstrations; for Track 2, where no external corpus exists, it caches a short factual description of each unseen target, prepends it to every tweet, and combines predictions and label-token scores from multiple LLMs via a Hungarian assignment algorithm aligned to the task's published class-count totals.

\begin{table*}[t]
\centering

\resizebox{\textwidth}{!}{%
\begin{tabular}{lllllcc}
\toprule
\textbf{Team} & \textbf{Paradigm} & \textbf{Core Model(s)} & \textbf{Key Methodological Features} & \textbf{Auxiliary Signals} & \textbf{Tr.~1} & \textbf{Tr.~2} \\
\midrule
TTLab~\cite{ttlab-2026} & Encoder Fine-tuning & BERT-based encoder & Cloze-style prompt fine-tuning with [MASK] mapping & None & \ding{51} & \ding{51} \\
Razan Alsulieman~\cite{alsulieman-2026} & Encoder Fine-tuning & MARBERT & LOTO validation under target shift, threshold tuning & Multi-task & \ding{51} & \ding{51} \\
Nabra~\cite{nabra-2026} & Encoder Fine-tuning & MARBERTv2 & Target-text sequence pairs, class weighting, back-translation & None & \ding{51} & -- \\
\midrule
SRK AI Lab~\cite{srk-2026} & Prompt-Based LLM & GPT-5.5, Gemini 3.6 Flash, Qwen 3.7 Plus & Independent multi-LLM classification with evidence-based auditing & Target evidence & \ding{51} & \ding{51} \\
JLShen~\cite{jlshen-2026} & Prompt-Based LLM & GPT-5.6, GPT-5.5, Claude-Opus, Claude-Sonnet & Cross-provider voting ensemble, confusion matrix derivation & Target cards & \ding{51} & \ding{51} \\
\textcolor{red}{HardStance}~\cite{faizaqamar-2026} & Prompt-Based LLM & Gemma-4-31B & Error-driven prompt calibration with target historical context & Context grounding & \ding{51} & -- \\
\midrule
NAMAA~\cite{namaa-2026} & Hybrid / RAG & ALLaM-7B (QLoRA), Arabic Encoders & Multi-task fine-tuning + retrieval-augmented in-context learning & Sentiment, Sarcasm & \ding{51} & \ding{51} \\
U4RASD~\cite{u4rasd-2026} & Hybrid / Prompting & Gemini 2.5 Flash, Arabic BERT, Claude & Few-shot prompting with targeted guidelines, self-consistency & Sentiment, Sarcasm & -- & \ding{51} \\
Orni~\cite{orni-2026} & Hybrid / Prompting & GPT-4o, MARBERTv2, AraBERT-Twitter & Few-shot structured reasoning vs. encoder ensemble comparison & Reasoning chains & \ding{51} & -- \\
\textcolor{red}{saliyah}~\cite{saliyah-2026} & Hybrid / Prompting & Fanar-9B, SILMA-9B, Jais-13B, ALLaM-7B, MARBERTv2 & Few-shot LLM voting vs. multi-task encoder comparison & Sentiment, Sarcasm & \ding{51} & \ding{51} \\
phonghoang~\cite{phonghoang-2026} & Hybrid / Augmentation & MARBERTv2, AraBERT, Gemini Flash & SWA multi-task ensemble + LLM pseudo-label retraining & ArabicStanceX & \ding{51} & -- \\
Axiom~\cite{axiom-2026} & Hybrid Cascade & MARBERTv2, Llama-3.1-8B-Instruct & High-confidence encoder gatekeeper with LLM fallback & Few-shot retrieval & \ding{51} & -- \\
\textcolor{red}{Meta\_Labs}~\cite{metalabs-2026} & Hybrid Routing & AraBERTv02-Twitter, Adversarial LLM & CaRPS: multi-task encoder routing to 4-call adversarial LLM & Sentiment, Sarcasm & \ding{51} & -- \\
\midrule
\textcolor{red}{rabia\_naz}~\cite{rabianaz-2026} & Fine-Tuned LLM & Qwen2.5-7B, ALLaM-7B & QLoRA adaptation + transductive agreement-based self-training & Sarcasm & \ding{51} & -- \\
ZHR~\cite{zhr-2026} & Fine-Tuned LLM / RAG & Gemma-4-31B-IT, ALLaM-7B, Qwen2.5-7B & Surface-matched synthetic demonstration retrieval + LoRA voting & Demonstrations & \ding{51} & \ding{51} \\
\midrule
Tarjih~\cite{tarjih-2026} & Auxiliary Reasoning & AraBERT, LLM generator/verifier & Generation of competing steelman arguments + verification scoring & Steelman args & \ding{51} & \ding{51} \\
Saigum~\cite{saigum-2026} & Auxiliary Reasoning & LLM prompting & Low-agreement filtering + AI discourse tag similarity retrieval & Discourse tags & \ding{51} & -- \\
Thakaa~\cite{thakaa-2026} & Auxiliary Reasoning & GPT-5.5, Gemini, Gemma & Target evidence grounding + Hungarian assignment class alignment & Target cards, ArabicStanceX & \ding{51} & \ding{51} \\
\bottomrule
\end{tabular}%
}
\caption{\textcolor{red}{System architecture and methodology synthesis across the 20 participating teams with accepted system papers.}}
\label{tab:system_synthesis}
\end{table*}

\paragraph{Key Findings Across Participating Systems.} Across both tracks, LLM-based systems dominate the top ranks while pure encoder fine-tuning tends to cluster lower. Track~1 leaders, JLShen's multi-LLM voting ensemble~\cite{jlshen-2026}, ZHR's retrieval-augmented classifier~\cite{zhr-2026}, and NAMAA's retrieval-augmented in-context learning~\cite{namaa-2026}, are echoed on Track~2 by ZHR, JLShen, and SRK AI Lab's multi-LLM auditing pipeline~\cite{srk-2026}, all built around LLM prompting, retrieval, or multi-model voting rather than encoder fine-tuning alone. Several teams confirm this pattern directly through their own ablations rather than leaderboard position alone: \textcolor{red}{saliyah} found few-shot LLMs consistently outperformed fine-tuned encoders across more than forty configurations~\cite{saliyah-2026}, and U4RASD found LLM prompting generalized to unseen targets better than any fine-tuned or ensemble-based alternative~\cite{u4rasd-2026}.


\subsection{Error Analysis and Failure Modes}\label{failure}
An examination of participant error logs and post-evaluation diagnostics highlights four main failure modes in Arabic cross-target stance detection:

\paragraph{1. The NONE-Class Dilemma and False Positive Stance Attribution.}
The most widespread failure across both tracks was the misclassification of neutral, informational, or irrelevant posts into active stance categories (\textit{Favor} or \textit{Against}). Because the primary competition metric $F_{avg2}$ excludes the \textit{None} class, models tuned to optimize $F_{avg2}$ frequently developed an aggressive bias toward predicting stance. News headlines, factual reports about governmental regulations, and rhetorical questions mentioning the target without an explicit personal attitude were systematically forced into polar labels. Teams that used balanced demonstrations or explicit break-even thresholding (e.g., JLShen~\cite{jlshen-2026}) achieved stronger all-class reliability ($F_{avg3}$) compared to those that relied on standard softmax classification.

\paragraph{2. Sarcasm, Irony, and Superficial Polarity Inversion.}
A substantial proportion of errors on \textit{Women Driving} and \textit{Trimester System} stemmed from ironic or sarcastic phrasing. Arabic social media users frequently deploy overt praise, celebratory religious supplications, or mock congratulations to convey intense opposition. Pure encoder models trained on sentiment signals routinely mapped positive affective vocabulary directly to \textit{Favor}, missing the ironic subtext. While multi-task learning with sarcasm supervision (e.g., in NAMAA~\cite{namaa-2026}, and \textcolor{red}{rabia\_naz}~\cite{rabianaz-2026}) mitigated this issue to some extent, resolving sophisticated cultural irony remained a key advantage for frontier LLMs prompted with explicit discourse guidelines.

\paragraph{3. Implicit Targets and Contextual Entailment.}
In numerous instances, the target entity was never explicitly mentioned in the tweet text. For example, tweets opposing \textit{E-Cars} often discussed power grid instability, battery replacement expenses, or the aesthetic preservation of traditional combustion engines without containing the keyword ``electric cars''. Similarly, tweets on \textit{Women Driving} frequently addressed general family transportation logistics, chauffeur licensing, or commuting safety. Systems that relied on simple target-post concatenation without external grounding or reasoning steps struggled to bridge the pragmatic gap between implicit arguments and the formal target label.

\paragraph{4. Dialectal Idioms and Orthographic Noise.}
Social media Arabic features substantial dialectal diversity (predominantly Gulf and Saudi vernacular in Mawqif-XT), non-standard orthography, repeated characters for emphasis, and localized colloquialisms. Encoder models pretrained primarily on Modern Standard Arabic (MSA) exhibited degraded representations on highly colloquial tweets. Back-translation was shown by Nabra~\cite{nabra-2026} to actively harm performance by corrupting dialectal idioms. Conversely, systems employing surface-matched demonstrations (ZHR~\cite{zhr-2026}) or native Arabic instruction-tuned LLMs (e.g., ALLaM and JAIS) showed superior resilience to dialectal variation.
\section{Conclusion}\label{sec:conclusion}
This paper presented an overview of StanceEval 2026, the second shared task on stance detection in Arabic social media text. The task evaluated cross-target generalization under two settings: Track1 focused on a thematically related target (\textit{Women Driving}), while Track2 evaluated unseen targets from different domains (\textit{E-Cars} and \textit{Trimester System}). StanceEval 2026 attracted 80 registrations from 12 countries and 30 participating teams, resulting in 21 and 13 officially ranked teams in Track1 and Track2, respectively, and 20 accepted system-description papers.

Participating teams explored diverse approaches, including fine-tuned language models, prompt-based LLMs, ensembles, multi-task learning, and retrieval-augmented systems. The best systems achieved \(F_{\mathrm{avg2}}\) scores of 0.8994 on Track1 and 0.9400 on Track2, substantially outperforming the strongest baselines. Counterintuitively, performance on unseen targets exceeded that on the related target.  Counterintuitively, performance on unseen targets exceeded that on the related target. \textcolor{red}{This difference may reflect factors such as stronger polarization on \textit{Trimester System} and more nuanced, sarcastic, and culturally complex discourse on \textit{Women Driving}; however, these factors may help explain the difference, but the current results do not confirm them as the main causes.} Future editions will expand the benchmark to broader Arabic dialects, implicit stance expressions, and multi-turn discourse.
\section*{Limitations }\label{sec:limit}

Several limitations should be considered when interpreting the StanceEval 2026 results. First, the comparison between Track1 and Track2 is not a controlled measure of the effect of target relatedness. The test targets differ in class distribution, degree of polarization, prevalence of neutral content, and linguistic complexity. For instance, \textit{Trimester System} is strongly skewed toward \textit{Against}, while \textit{Women Driving} has a nearly balanced distribution of \textit{Favor} and \textit{Against} instances and contains more nuanced stance expressions. Thus, the higher performance observed on Track~2 may partly reflect differences in target difficulty rather than stronger generalization to thematically unseen targets.

Second, the primary ranking metric, $F_{avg2}$, considers only the \textit{Favor} and \textit{Against} classes. This is important because distinguishing \textit{None} from explicit stance emerged as a common source of error, and the prevalence of this class varies considerably across targets. To provide a broader assessment, we additionally report $F_{avg3}$ and accuracy.

Third, the evaluation includes a limited number of test targets and a relatively small Mawqif-XT test set. The data also contain substantial Gulf and Saudi dialectal content and culturally grounded phenomena, including sarcasm, implicit stance, and socio-cultural argumentation. Consequently, the observed performance patterns may not generalize to other topics, domains, or Arabic varieties.

Finally, the qualitative failure modes discussed in Section~\ref{failure} are derived from post-evaluation examination of system outputs and participant analyses rather than a systematically annotated error study. They should therefore be interpreted as plausible explanations for observed performance patterns rather than definitive causal conclusions.

\section*{Acknowledgments}
The StanceEval 2026 organizers would like to acknowledge the support received from the Saudi Data and AI Authority (SDAIA) and King Fahd University of Petroleum and Minerals (KFUPM) through the SDAIA-KFUPM Joint Research Center for Artificial Intelligence. 

\normalem
\bibliography{custom}

\appendix
\appendix
\section{\textcolor{red}{Dataset Examples}}\label{app:examples}

\textcolor{red}{Tables~\ref{tab:aligned_examples} and~\ref{tab:nonaligned_examples}
provide representative examples from Mawqif-XT.}

\begin{table*}[t]
{\color{red}
\centering
\small
\setlength{\tabcolsep}{4pt}
\renewcommand{\arraystretch}{1.2}

\begin{tabularx}{\textwidth}{
>{\raggedright\arraybackslash}p{1.7cm}
>{\raggedright\arraybackslash}X
>{\centering\arraybackslash}p{1.0cm}
>{\centering\arraybackslash}p{1.1cm}
>{\centering\arraybackslash}p{0.9cm}
}
\toprule
\textbf{Target} & \textbf{Text} & \textbf{Stance} & \textbf{Sentiment} & \textbf{Sarcasm} \\
\midrule

Women Driving &
\RL{البنت السعوديه قدها والى غيرها مو احسن منها سواء بالقياده او بالحشمه او بالعفاف اوبالحياء او بالحجاب . وكل ماهو بالبنت السعوديه يمثل البنت المسلمه. كلنا معك. \#المراه\_السعوديه\_تسوق} &
Favor &
Positive &
No \\

\midrule

Women Driving &
\RL{\#لن\_تقودي والنهاية يعني من قالكم اكبر همنا السيارات والقيادة .. احنا نحتاج مترو او باصات تساعدنا او تسهل علينا المشاوير من دون سواق وتكاليف} &
Against &
Negative &
Yes \\

\midrule

Ecars &
\RL{السيّارات الكهربائية أفضل مئات المرات في توزيع العزم بين الإطارات من السيارات التقليدية} &
Favor &
Positive &
No \\

\midrule

Ecars &
\textnormal{MENTION} \RL{ما زالت مشكلة السيارات الكهربائية هي مدى الشحنة جدًا تعبانة} &
Against &
Negative &
No \\

\midrule

Trimester &
\RL{كنت وما أزال أرى أن تقسيم الدراسة إلى ثلاثة فصول أفصل من فصلين الاستمتاع بالاجواء الشتوية والصيفية تحريك السياحة الداخلية محاولة لتحريك جمود الاقتصاد إراحة الطلبة والمعلمين إمكانية التعويض في الدرجات \LR{URL}} &
Favor &
Positive &
No \\

\midrule

Trimester &\RL{\LR{MENTION} فعلا عدد الحصص اصبح اكثر من عدد الدروس وهذا من سلبيات الفصول الثلاثه} &
Against &
Negative &
No \\

\bottomrule
\end{tabularx}
}
\caption{Representative examples of aligned stance--sentiment polarity in Mawqif-XT.}
\label{tab:aligned_examples}
\end{table*}

\begin{table*}[t]
{\color{red}
\centering
\small
\setlength{\tabcolsep}{4pt}
\renewcommand{\arraystretch}{1.2}

\begin{tabularx}{\textwidth}{
>{\raggedright\arraybackslash}p{1.7cm}
>{\raggedright\arraybackslash}X
>{\centering\arraybackslash}p{1.0cm}
>{\centering\arraybackslash}p{1.1cm}
>{\centering\arraybackslash}p{0.9cm}
}
\toprule
\textbf{Target} &
\textbf{Text} &
\textbf{Stance} &
\textbf{Sentiment} &
\textbf{Sarcasm} \\
\midrule

Women Driving &
\RL{\#لن\_تقودي انا سائق لأمي وأختي وبنتي وزوجي وأفتخر بشرف رعايتك وحمايتك لا تغرنك هتافاتهم هم يتربصون لا تضعفي أعلنيها \#لا\_للمتربصين\_بالمسلمات} &
Against &
Positive &
No \\

\midrule

Ecars &
\RL{الرؤية غائبة والهدف متواضع والتنفيذ يتعثر. تصميم وتصنيع سيارة كهربية مصرية شيء يمكن تحقيقه يقينا لو اخلصنا النية واعتمدنا بشفافية على قدراتنا وبنفس تكلفة اللجان والاستشاري الالماني والشريك الصيني ولجانكم الحكومية وزير قطاع الأعمال قريبًا سيارة كهربائية \LR{URL}} &
Favor &
Negative &
No \\

\midrule

Trimester &
\RL{\LR{MENTION} أملنا في الله ثم في ولاة أمرنا أنهم ينظرون في وضع الفصل الدراسي الثالث ويلغونه أن شاءالله والاكتفاء بفصلين دراسيين فقط حسب المعتاد} &
Against &
Positive &
No \\

\bottomrule
\end{tabularx}
}
\caption{Examples of non-aligned stance--sentiment polarity in Mawqif-XT.}
\label{tab:nonaligned_examples}
\end{table*}

\section{\textcolor{red}{Zero-shot LLM Prompt}}
\label{app:llm_prompt}

\textcolor{red}{The following prompt was used for all zero-shot LLM baselines:}

\begin{quote}
\textcolor{red}{You are a stance classification model. Classify the stance of the tweet toward the given target. Use exactly one label only from: Favor, Against, None. Favor means the tweet supports the target. Against means the tweet opposes the target. None means the tweet is neutral, unrelated, or expresses no clear stance toward the target. Output exactly one label only and nothing else.}

\textcolor{red}{\texttt{Target: \{target\}}}\\
\textcolor{red}{\texttt{Tweet: \{text\}}}
\end{quote}

\section{Additional System Descriptions}\label{app:systems}
This appendix details the remaining participating systems, summarized in Table~\ref{tab:system_synthesis} but not discussed individually in Section~\ref{subsec:system_descriptions}.

\textbf{Encoder-Based Fine-Tuning Methods.} 
TTLab~\cite{ttlab-2026}: The paper introduces a system to address the StanceEval2026 shared task. The system fine-tunes a BERT-based encoder on prompts in a cloze-style format, where the target class is masked with the token [MASK]. The training objective is to map this mask token to a set of 3 classes: Favor, Against, and None.

Nabra~\cite{nabra-2026} participated in Track 1 and compared seven configurations built primarily around MARBERTv2 using a TARGET [SEP] TWEET input formulation. The experiments covered standard fine-tuning, class weighting, dropout with learning-rate scheduling, back-translation, oversampling, undersampling, and zero-shot inference with Qwen2.5-7B. The simple baseline outperformed methods designed to improve the minority NONE class, which increased Favg3 but reduced performance on the Favor and Against classes. Back-translation also degraded performance, which the authors attribute to noise introduced when translating dialectal Arabic using models primarily suited to MSA.

\textbf{Prompt-Based LLM Methods}

\textcolor{red}{HardStance}~\cite{faizaqamar-2026} compared three fine-tuned AraBERT variants with zero-shot generative LLMs for Track 1. Their final system used Gemma-4-31B with a prompt refined through development-set error analysis. The prompt included instructions for interpreting dialectal and implicit expressions, with brief historical context about Women Driving.

\textbf{Hybrid Encoder–LLM Methods}

U4RASD~\cite{u4rasd-2026} participated in Track 2 for unseen-target Arabic stance detection and conducted a broad comparison of approaches, including Arabic BERT encoders, focal and contrastive losses, sentiment and sarcasm multitask learning, GPT-4o-mini zero-shot prompting, self-consistency with Claude escalation, Gemini 2.5 Flash few-shot prompting, LoRA fine-tuning of several LLMs, and model ensembling. Gemini 2.5 Flash few-shot prompting generalized best to the unseen target, outperforming the fine-tuned and ensemble alternatives, using an extended prompt with eight labeled examples and explicit instructions distinguishing stance from sentiment and implicit targets, sarcasm, and dialectal variation.

Orni team~\cite{orni-2026} participated in Track 1 using two main approaches: fine-tuning-based models and GPT-4o with prompt engineering. The fine-tuning approach primarily combines MARBERTv2 and AraBERT-Twitter, while also exploring QLoRA-based LLM adaptation and other strategies to improve generalization. The second approach uses GPT-4o without fine-tuning, relying on few-shot demonstrations, structured reasoning, target-specific context, and self-consistency voting. The prompted GPT-4o system outperformed the fine-tuning-based systems.

saliyah~\cite{saliyah-2026} compares fine-tuned Arabic encoders with few-shot Arabic LLMs for both tracks. The encoder models includes MARBERTv2, AraBERT-large, AraELECTRA, and CAMeLBERT-sentiment, trained with multi-task learning using sentiment, sarcasm, and rationale annotations, while the LLMs, Fanar-9B, SILMA-9B, Jais-13B, and ALLaM-7B, are used without fine-tuning in a fixed 3-shot setting. The paper also tests chain-of-thought, self-consistency, and guidance-rich prompting. For Track 1, the final system combines two LLMs through a simple voting rule. Overall, the few-shot LLMs outperformed the fine-tuned encoders, and simpler prompting methods performed better than more complex inference strategies.

Phonghoang team~\cite{phonghoang-2026} followed a two-stage approach. They first trained a MARBERTv2 and AraBERT ensemble using multi-task learning, cross-validation, R-Drop regularization, and Stochastic Weight Averaging (SWA). They then used Gemini Flash to generate pseudo-labels for the 352 Women Driving test instances. These were combined with external Women Driving data from the ArabicStanceX dataset~\cite{alkhathlan-2025-arabicstancex} and high-confidence Women Empowerment examples to retrain a MARBERTv2 ensemble.

Axiom~\cite{axiom-2026} used MARBERTv2 as a gatekeeper for high-confidence case predictions, while lower-confidence predictions were routed to Llama-3.1-8B-Instruct as a fallback for harder cases. The LLM was provided with balanced few-shot examples retrieved from the training data. This hybrid approach improved over the standalone MARBERTv2 baseline.

The \textcolor{red}{Meta\_Labs} team~\cite{metalabs-2026} proposed Calibrated Routing and Pseudo-Label Self-Training (CaRPS). They fine-tuned Arabertv02-twitter for stance detection, with sentiment and sarcasm as auxiliary tasks. Their original design used the encoder for the initial prediction and routed uncertain cases to a four-call adversarial LLM resolver; however, this routing underperformed using the adversarial LLM directly, so the team used the LLM alone for the official test set. A separate development experiment in which LLM-generated labels for Women Empowerment were used to retrain the encoder improved its performance over the baseline.

\textbf{Fine-tuned LLM Ensembles}

Rabia\_naz~\cite{rabianaz-2026} fine-tunes three open LLMs (Qwen2.5-7B, ALLaM-7B, and a sarcasm-aware Qwen variant) with QLoRA on the official Mawqif-v2 training data. It adapts them to the related target "Women Driving" via transductive self-training, where unlabeled test tweets join the training pool only if multiple independent models agree on their pseudo-label. The three fine-tuned models combine via majority vote, with the two Qwen-based members contributing one-hot labels and ALLaM breaking ties using its softmax probabilities. Sarcasm-annotation supervision improved the ensemble's performance on Track 1.

\textbf{LLM-Generated Reasoning as Auxiliary Signal}

Tarjih~\cite{tarjih-2026} reframes Arabic stance detection as a verification problem. For each tweet, an LLM generates a steelman argument for each stance label, then a second LLM (the verifier) assesses how well each argument matches the tweet. These argument-tweet pairs are jointly encoded with AraBERT and fed into one of three prediction heads, a gated classifier, a pair scorer, and a ranking read-out, to produce the final stance label. This lets the system reason over competing interpretations rather than committing upfront. The pair scorer performed best on the seen-target track, and the gated classifier performed best on the unseen-target track.

\section{Full Leaderboards} \label{app:leaderboards}
Tables~\ref{tab:results_track_1} and~\ref{tab:results_track_2} list the complete official rankings for Track 1 (21 teams) and Track 2 (13 teams) on the evaluation set. Systems are ranked by $F_{avg2}$; $F_{avg3}$ and overall accuracy are reported alongside for completeness.

\begin{table}[h]
\centering
\small

\begin{tabular}{l r r r}
\toprule
Team & $F_{avg2}$ & $F_{avg3}$ & Acc. \\
\midrule
JLShen & 0.8994 & 0.6646 & 0.8580 \\
ZHR & 0.8960 & 0.6862 & 0.8551 \\
NAMAA & 0.8719 & 0.6722 & 0.8153 \\
Saigum & 0.8685 & 0.6457 & 0.8210 \\
HardStance & 0.8661 & 0.6746 & 0.8267 \\
Thakaa & 0.8522 & 0.6792 & 0.8125 \\
phonghoang & 0.8431 & 0.5621 & 0.8011 \\
SRK AI Lab & 0.8420 & 0.6229 & 0.7813 \\
\textcolor{red}{Meta\_Labs} & 0.8289 & 0.6267 & 0.7898 \\
AH-Translit & 0.8279 & 0.6126 & 0.7869 \\
Orni & 0.8270 & 0.6297 & 0.7841 \\
\textcolor{red}{rabia\_naz} & 0.7564 & 0.5233 & 0.7216 \\
\textcolor{red}{saliyah} & 0.7469 & 0.5480 & 0.7131 \\
Razan Alsulieman & 0.7435 & 0.5253 & 0.7017 \\
Tarjih & 0.7315 & 0.5057 & 0.6960 \\
L3IA & 0.7208 & 0.5089 & 0.6790 \\
\textcolor{red}{TTLab} & 0.7136 & 0.5313 & 0.6761 \\
Axiom & 0.6910 & 0.5032 & 0.6534 \\
Cristiano Ronaldo & 0.6872 & 0.4581 & 0.6563 \\
Nabra & 0.6704 & 0.4591 & 0.6222 \\
nour\_mubarak & 0.6625 & 0.4854 & 0.6165 \\
\bottomrule
\end{tabular}
\caption{\textcolor{red}{Full results for Track 1 on the Evaluation set, ranked by $F_{avg2}$.}}
\label{tab:results_track_1}
\end{table}

\begin{table}[h]
\centering
\small

\begin{tabular}{l r r r}
\toprule
Team & $F_{avg2}$ & $F_{avg3}$ & Acc. \\
\midrule
ZHR & 0.9400 & 0.8792 & 0.9224 \\
JLShen & 0.9357 & 0.8670 & 0.9130 \\
SRK AI Lab & 0.9192 & 0.8374 & 0.8882 \\
Thakaa & 0.9178 & 0.8353 & 0.8882 \\
AH-Translit & 0.9125 & 0.8360 & 0.8820 \\
HardStance & 0.9086 & 0.8268 & 0.8742 \\
NAMAA & 0.9016 & 0.8254 & 0.8680 \\
U4RASD & 0.8741 & 0.7694 & 0.8307 \\
\textcolor{red}{saliyah} & 0.8342 & 0.6770 & 0.7826 \\
Razan Alsulieman & 0.8100 & 0.6859 & 0.7578 \\
Tarjih & 0.7699 & 0.6370 & 0.7252 \\
\textcolor{red}{TTLab} & 0.7414 & 0.6564 & 0.7096 \\
nour\_mubarak & 0.7246 & 0.6023 & 0.6801 \\
\bottomrule
\end{tabular}
\caption{\textcolor{red}{Full results for Track 2 on the Evaluation set, ranked by $F_{avg2}$.}}
\label{tab:results_track_2}
\end{table}

\end{document}